\documentclass[10pt,twocolumn,letterpaper]{article}

\usepackage[pagenumbers]{cvpr}
\usepackage[T1]{fontenc}
\usepackage{multirow}
\usepackage{makecell}
\usepackage{hhline}

\usepackage[most]{tcolorbox}
\tcbset{
  appendixprompt/.style={
    colback=gray!5,
    colframe=gray!60!black,
    boxrule=0.6pt,
    fonttitle=\bfseries,
    left=0.6em,
    right=0.6em,
    top=0.55em,
    bottom=0.55em,
    width=0.96\textwidth,
    arc=4pt
  }
}

\definecolor{cvprblue}{rgb}{0.21,0.49,0.74}
\usepackage[breaklinks,colorlinks,allcolors=cvprblue]{hyperref}
\hypersetup{
  pdftitle={TryOnReward: Learning Foveated Consistency for Reinforcement Fine-Tuning of Virtual Try-On},
  pdfauthor={Xueheng Li, Yong Liu, Xiaolong Fu, Wen Xue, Chengjun Xie, Yipeng Sun, Yan Li, Simiu Gu}
}

\title{TryOnReward: Learning Foveated Consistency for Reinforcement \\
Fine-Tuning of Virtual Try-On}
\author{%
  Xueheng Li\textsuperscript{1,2}\thanks{These authors contributed equally.},
  Yong Liu\textsuperscript{1*},
  Xiaolong Fu\textsuperscript{1}, Wen Xue\textsuperscript{1,3},\\
  Chengjun Xie\textsuperscript{2},
  Yipeng Sun\textsuperscript{1}\thanks{Corresponding author.},
  Yan Li\textsuperscript{1}, Simiu Gu\textsuperscript{1}\\[0.5em]
  \textsuperscript{1}JD.com\\
  \textsuperscript{2}University of Science and Technology of China\\
  \textsuperscript{3}South China University of Technology
}

\date{}

\makeatletter
\patchcmd{\@maketitle}{\vspace*{24pt}}{\vspace*{10pt}}{}{%
  \PackageError{tryon-arxiv}{Could not adjust title spacing}{}%
}
\makeatother

\begin{document}
\raggedbottom
\maketitle
\begin{abstract}
Virtual Try-On (VTON) aims to dress a person with the reference garment, producing visually reasonable results aligned with human preferences. Turning this preference-oriented goal into an actionable objective relies on a scoring function aligned with human taste. However, classic fidelity metrics exhibit weak correlation with human judgments, and generic VLMs fail to provide the discriminative granularity demanded by try-on quality evaluation, which hinges on faithfully preserving garment and person details. This shortcoming is further exacerbated in the reinforcement fine-tuning (RFT) optimization and leads to severe reward hacking. To this end, we present TryOnReward, a fine-grained reward model tailored for VTON. Built on a vision-language backbone, it adopts a foveation calibration objective that grounds each quality dimension in the relevant region to avoid global shortcut learning. Meanwhile, TryOnReward jointly optimizes pairwise preferences and per-dimension quality scores via margin-aware supervision, leveraging both relative and absolute quality signals. For model training and evaluation, we build TryOnRank-100K, a human-annotated per-dimension rating dataset, alongside TryOn-Bench and TryOnRewardBench --- two benchmarks covering diverse real scenarios. Extensive experiments confirm that our proposed TryOnReward significantly outperforms generic judges in human preference alignment, and when serving as the RFT reward function, it consistently yields human-preferred try-on results across multiple baselines.
\end{abstract}

\section{Introduction}
\begin{figure}[t]
    \centering
    \includegraphics[width=\columnwidth]{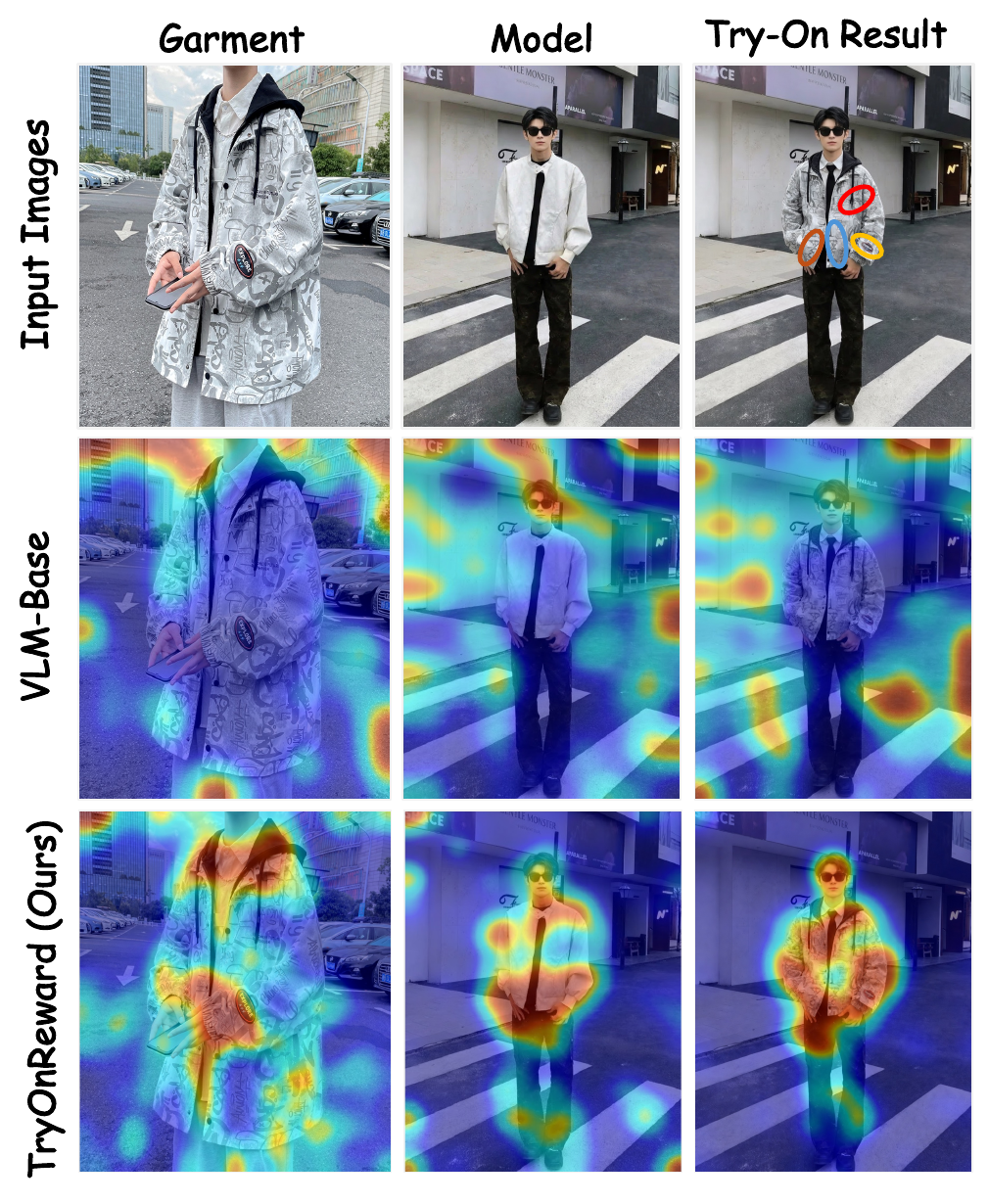}
    
    \caption{Visualization of the regions attended by different reward models. Generic VLMs tend to produce diffuse attention detrimental to fine-grained recognition tasks. As a comparison, our TryOnReward correctly attends to the critical regions required for foveated consistency assessment.}
    \label{fig:motivation}
    
\end{figure}

Virtual try-on (VTON) seeks to digitally dress a target person in a specified garment~\cite{chong2025catvton, feng2026omnitry, liu2026oxygen}. Owing to modern generative models, this task has advanced at a remarkable pace. As synthesis quality matures, the frontier is shifting from producing a \emph{plausible} result to producing a \emph{preferred} one. Reaching this frontier hinges on a scoring function aligned with human taste, which plays two complementary roles: it benchmarks try-on models against human judgment, and it supplies
  the optimization signal for reinforcement fine-tuning (RFT)~\cite{zheng2025diffusionnft, liu2026flow, xu2026qwen}, an increasingly dominant post-training paradigm that steers policies toward
  human-preferred outputs~\cite{wang2025unified, luo2025editscore, zhong2025comprehensive}. A reward model that reliably captures human preference is therefore central to screening high-quality generations and advancing virtual try-on systems.

  Scoring try-on quality, however, is uniquely demanding. Prior studies~\cite{wei2026vtonqa, wu2026high} highlight garment fidelity, preservation of person-specific attributes, and visual realism as central criteria for assessing try-on quality. Unlike general image generation, VTON must render a realistic result while faithfully preserving fine-grained garment attributes (\textit{e.g.}, logo, texture, and weave) and person identity, free of silhouette distortion or editing artifacts. Quality therefore turns on region-level details that a single holistic score tends to overlook. Yet existing scorers often fail to capture such region-level details: classic fidelity metrics (\textit{e.g.}, SSIM, LPIPS, FID) correlate poorly with human judgment, while general-purpose reward models---ImageReward~\cite{xu2023imagereward}, BaseReward~\cite{zhang2025basereward}, HPSv3~\cite{ma2025hpsv3}, and EditReward~\cite{wu2025editreward}---primarily target broad image-text alignment, instruction following, and overall perceptual quality. Their scoring dimensions do not directly match these VTON-specific requirements, limiting their direct applicability to fine-grained try-on evaluation. Off-the-shelf open-source VLMs~\cite{bai2025qwen3, wang2025internvl3, chng2025sensenova, gao2026videosearcher, li2026qprobe}  and proprietary models~\cite{openai_gpt54} used as judges likewise lack the fine-grained discrimination that try-on demands, missing subtle but decisive failures such as garment-detail loss and identity drift. This blind spot is not merely an evaluation nuisance---under RFT it is actively exploited: when a holistic reward drives RFT, the policy drifts toward globally attractive yet garment-mismatched or identity-shifted images while the measured reward keeps climbing.

  Two design gaps explain this weakness. \emph{First, existing reward models underuse visual evidence.} Fine-grained VTON evaluation requires comparing specific regions across images---the reference garment
  against the output's clothing region, and the person's identity-bearing regions between input and output---yet VLMs often lean on language priors and coarse semantic representations rather than
  task-relevant visual tokens~\cite{guan2024hallusionbench, zheng2025mllms}, so their attention is easily drawn to salient but non-informative content~\cite{lin2025boosting}, as shown in Fig. ~\ref{fig:motivation}.
  \emph{Second, they represent
  reward too narrowly.} Pairwise preference learning~\cite{ma2025hpsv3, zhang2025basereward} ranks preferred candidates above rejected ones---well suited to ranking and RFT, but capturing only preference
  direction, so pairs rated $(5,4)$ and $(5,1)$ are treated identically despite very different quality gaps; pointwise regression~\cite{li2026q, zhang2026vq} conveys absolute quality but is sensitive to
  annotation subjectivity and enforces no explicit ordering. The two are complementary, motivating a unified formulation that captures both relative ranking and absolute score semantics.

  To address these challenges, we present TryOnReward, a fine-grained reward model tailored to VTON. TryOnReward serves both as a human-aligned evaluator of try-on quality and as a reward function for optimizing VTON policies. Guided by these task requirements, we model human preference along three core dimensions---Garment Consistency, Identity Consistency, and Visual Quality---and learn task-specific criteria for each using a vision-language backbone. The model combines mask-guided foveation calibration with joint pairwise--pointwise supervision, grounding its judgments in relevant visual evidence while capturing preference ordering, rating semantics, and preference strength. To train and evaluate the model, we build TryOnRank-100K, a large-scale dataset of 99,286 human-annotated preference pairs with per-dimension ratings, and TryOnRewardBench, a benchmark spanning diverse input formats, person distributions, and real-world scenarios. Extensive experiments show that TryOnReward achieves substantially stronger alignment with human preferences than general-purpose judges and, when used as the reward for RFT, improves the quality of generated try-on results across different policy models.

  The contributions of this paper are summarized as follows:
  \begin{itemize}
      \item We present TryOnReward, a VTON-specific reward model that decomposes human preference into garment consistency, identity consistency, and visual quality, and we build TryOnRank-100K and
  TryOnRewardBench to support training and systematic evaluation of try-on reward models.
      \item We introduce a foveation calibration objective that grounds each quality dimension in its task-relevant region, anchoring reward predictions in reliable visual evidence; and a unified
  reward-learning formulation that couples pairwise preference learning with pointwise score regression to jointly capture preference ordering, rating semantics, and preference strength.
      \item Extensive experiments on TryOnRewardBench, DressCode, VITON-HD, and TryOnBench demonstrate that TryOnReward aligns with human preference far better than general judges and, as an RFT reward,
  consistently improves try-on quality across baselines.
  \end{itemize}

\begin{figure*}[t]
    \centering
    \includegraphics[width=\textwidth]{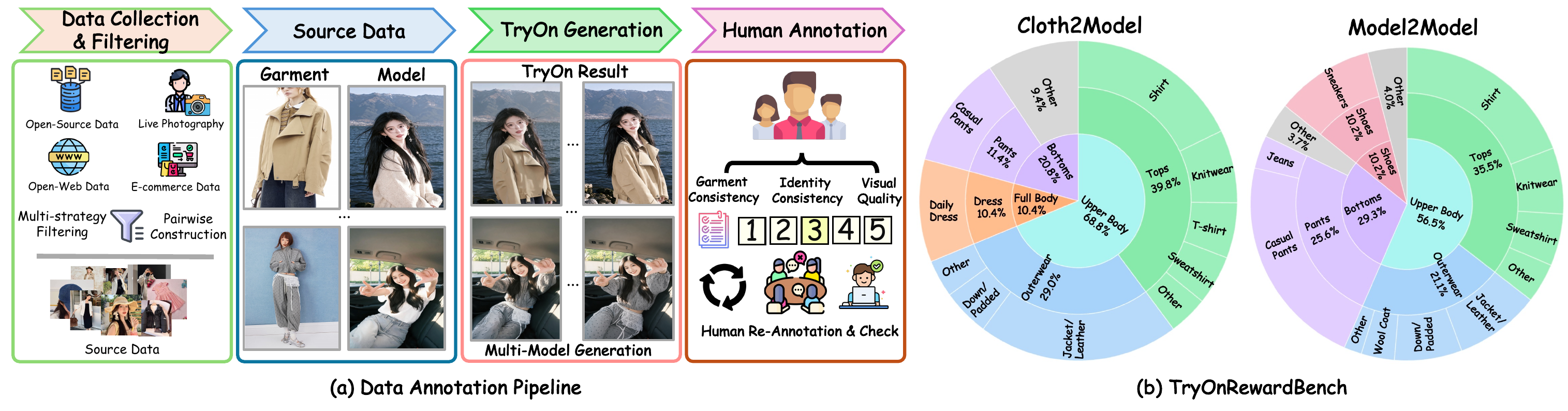}   
    
    \caption{Overview of the data construction pipeline and TryOnRewardBench composition. (a) Garment and person images are collected and filtered for try-on generation, followed by independent annotation and cross-checks by two teams to obtain scores and preference labels. (b) The benchmark includes Cloth2Model (flat-lay references) and Model2Model (worn references), covering major categories such as upper-body garments, bottoms, and shoes.}
    \label{fig:data_benchmark}
    
\end{figure*}

\section{Related Work}
\textbf{Virtual Try-On:}
Recent advances in VTON have been largely driven by diffusion models, with methods such as TryOnDiffusion~\cite{zhu2023tryondiffusion}, OOTDiffusion~\cite{xu2025ootdiffusion}, and CatVTON~\cite{chong2025catvton} improving garment fidelity through effective garment feature integration. More recent approaches further explore Diffusion Transformers and unified try-on frameworks, including Leffa~\cite{zhou2025learning} and OmniTry~\cite{feng2026omnitry}, targeting finer detail preservation, broader applicability, and mask-free generation. Meanwhile, reinforcement fine-tuning (RFT)-based post-training has been introduced into VTON, with Tstars-Tryon~\cite{chen2026tstars} leveraging multi-reward optimization to improve generation quality. However, existing VTON methods still lack a dedicated fine-grained reward model capable of evaluating task-specific consistency and providing reliable supervision for RFT-based optimization.

\noindent \textbf{Reward Model:}
Reward models have become a key component of RFT-based post-training for generative models by providing human-aligned optimization signals. In visual generation, methods such as the HPS series~\cite{wu2023human, ma2025hpsv3} learn human preferences from large-scale annotations, while recent VLM-based approaches further extend reward modeling to broader scenarios. RewardDance~\cite{wu2025rewarddance} explores scalable VLM-based reward prediction, OneReward~\cite{gong2025onereward} develops a unified reward model for multi-task image generation, and EditReward~\cite{wu2025editreward} targets instruction-guided image editing. Nevertheless, these general-purpose reward models are primarily designed for single-image generation or editing and are insufficient for VTON, which requires multi-input reasoning and region-level evaluation of garment and identity consistency. In contrast, we introduce a VTON-specific reward model that decomposes human preferences into garment consistency, identity consistency, and visual quality, and provides fine-grained task-aware supervision through unified pairwise--pointwise learning with Foveation Calibration Objective.

\begin{figure*}[t]
    \centering
    \includegraphics[width=\textwidth]{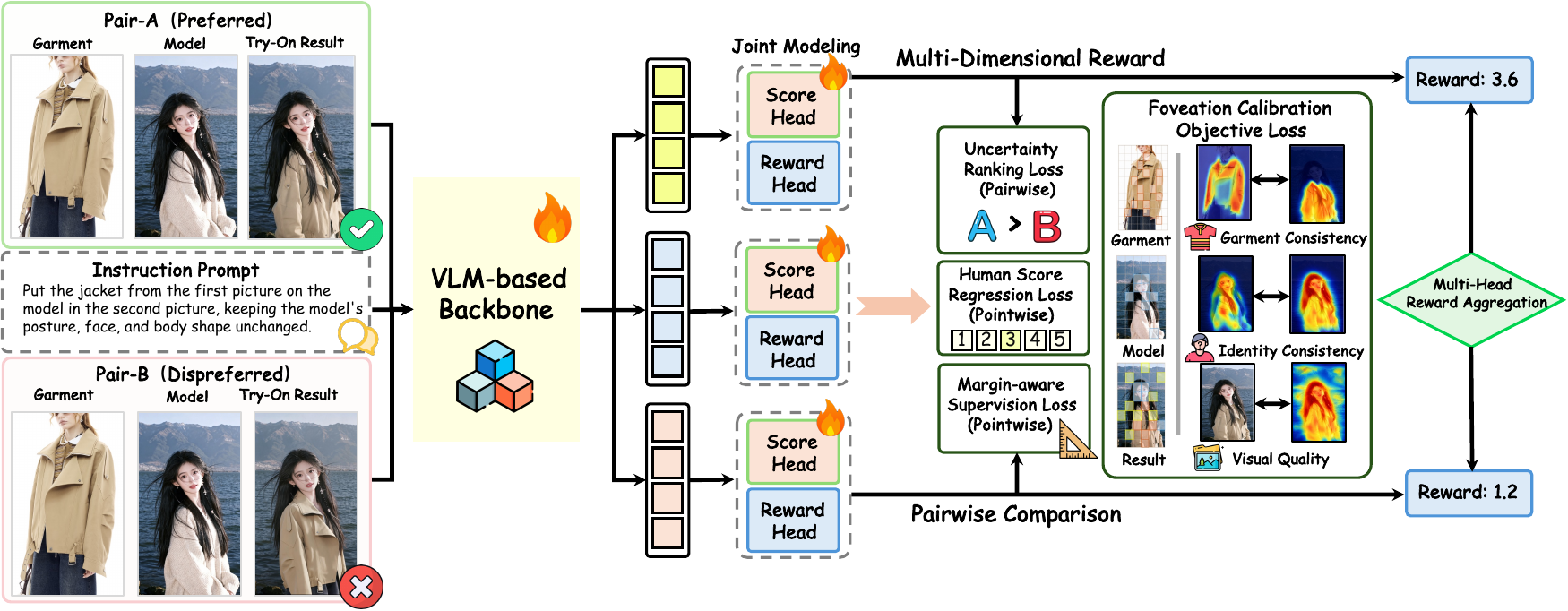}   
    \caption{Overview of TryOnReward. Built upon a shared VLM-based backbone, it performs dimension-specific evaluation for garment consistency, identity consistency, and visual quality. TryOnReward jointly models pairwise preferences and point-wise human scores for garment consistency, identity consistency, and visual quality, while leveraging mask-guided Foveation Calibration Objective to ground reward prediction in dimension-specific visual evidence.}
    \label{fig:framework}
\end{figure*}

\section{Dataset Curation and Construction}
\textbf{TryOnRank-100K:} 
To train TryOnReward, we construct TryOnRank-100K using a four-stage pipeline comprising source data collection and filtering, candidate generation, pair construction, and preference annotation, as illustrated in Fig.~\ref{fig:data_benchmark}. We collect and curate diverse garment and person images from open-source datasets~\cite{morelli2022dress, choi2021viton}, public web sources, e-commerce platforms, and live photography. Using the resulting garment-person conditions, we generate diverse try-on candidates with four advanced image editing models: Nano Banana Pro~\cite{nanopro}, Qwen-Image-Edit-2511~\cite{wu2025qwen}, FireRed-Image-Edit 1.0~\cite{team2026firered}, and JoyImage~\cite{song2026joyai}. After candidate-level quality filtering, we retain 87,871 instances, each consisting of a garment image, a person image, a generated result, and the corresponding instruction. Candidates under the same garment-person condition are then paired, yielding 118,074 raw pairs for preference annotation.

All generating-model identities are concealed from annotators under a blind evaluation protocol. Two separate annotation teams independently evaluate each candidate using the same criteria for three dimensions: Garment Consistency (garment details, texture, color, and fit), Identity Consistency (face, body shape, limbs, and pose), and Visual Quality (sharpness, noise, blur, and artifacts).
Within each team, three distinct annotators score each dimension on a 1--5 scale, yielding nine annotators per candidate per team. The mean of each team's three ratings per dimension serves as a provisional score label and determines the relative preference between candidates A and B in each pair.
We cross-check the two teams' preference labels and discard pairs with opposite preference orderings. For retained pairs, we average all six ratings per candidate and dimension across the two teams and round the mean to the nearest integer to obtain the final score label; the final pairwise preference labels are derived from these scores.
This procedure yields 99,286 high-quality preference pairs, which constitute the final TryOnRank-100K training set.

\noindent  \textbf{TryOn-Bench and TryOnRewardBench:}
To comprehensively evaluate the reward model and the try-on models trained with it, we construct TryOn-Bench and TryOnRewardBench. They are built from the same garment--person condition images; the difference is that TryOnRewardBench additionally incorporates generations from multiple models along with human preference annotations on these results, in order to directly assess whether TryOnReward's recognition of different outputs aligns with real human preferences.
Specifically, we generate candidate results using six models: Nano Banana Pro~\cite{nanopro}, SeedDream 5.0~\cite{bytedance2026seedream}, SeedDream 4.5~\cite{bytedanceseed2025seedream45}, Qwen-Image-Edit-2511~\cite{wu2025qwen}, JoyImage~\cite{song2026joyai} and GPT-Image-2~\cite{gptimage2_model_card}. TryOnRewardBench includes outputs from generators not used to construct TryOnRank-100K, allowing us to assess generalization to unseen generator output distributions.
The benchmarks are designed to cover two complementary VTON settings: Cloth2Model and Model2Model. Cloth2Model contains 9,844 pairs and uses flat-lay product images as garment references, corresponding to the standard product-to-person try-on scenario. Model2Model comprises 7,045 pairs in which the source garments are depicted on another person. 
Compared with Cloth2Model, it is more challenging because the model must disentangle garment appearance from the source wearer and perform complex cross-person appearance transfer. The garment-category distributions of the two subsets are presented in Fig.~\ref{fig:data_benchmark}(b). It is noted that TryOnRank-100K and TryOnRewardBench are strictly disjoint, with no shared source images, conditions, or generated samples.

\section{Methods}
Given a reference garment image \(I^g\), a target person image \(I^p\), and a generated try-on result \(I^c\), our goal is to learn fine-grained rewards along three dimensions: Garment Consistency, Identity Consistency, and Visual Quality. We denote the dimension set by \(\mathcal{D}=\{\mathrm{gar},\mathrm{id},\mathrm{vis}\}\). Each training sample contains a garment-person condition \((I^g,I^p)\) and a pair of try-on candidates \(\mathcal{C}=\{A,B\}\). For each \(c\in\mathcal{C}\) and \(d\in\mathcal{D}\), a human rating \(s_c^d\in\{1,\ldots,5\}\) is provided. These annotations supervise relative preference ordering, absolute rating semantics, and preference strength.

\subsection{Model Architecture}
As shown in Figure \ref{fig:framework}, TryOnReward is built upon a shared vision-language backbone.
With a reference garment \(I^g\), a target person \(I^p\), and a generated try-on result \(I^c\), the model extracts a dimension-conditioned multimodal representation \(\mathbf{h}_c^d=F_\theta(I^g,I^p,I^c,P^d)\) for each evaluation dimension \(d\in\mathcal{D}\), where \(P^d\) denotes the corresponding dimension-specific evaluation prompt. To disentangle the three evaluation criteria while retaining a shared visual representation, we attach an independent reward head \(R_d\) and score head \(S_d\) to each dimension:
\begin{equation}
(\mu_c^d,(\sigma_c^d)^2)=R_d(\mathbf{h}_c^d),
\quad
\mathbf{z}_c^d=S_d(\mathbf{h}_c^d)\in\mathbb{R}^{5}.
\label{eq:head}
\end{equation}
The two heads capture complementary aspects of human supervision. Fine-grained VTON judgments can be ambiguous, particularly when two candidates differ only subtly. Following uncertainty-aware reward modeling \cite{ma2025hpsv3,wu2025editreward}, the reward head parameterizes a Gaussian distribution \(p_c^d(r)=\mathcal{N}(r;\mu_c^d,(\sigma_c^d)^2)\) and is used to provide supervisory signals for pairwise preferences. Its mean represents the scalar reward, while its standard deviation captures prediction uncertainty. In parallel, the score head predicts logits over the five human rating levels, providing explicit pointwise supervision. 
The joint optimization of the two heads allows the model to simultaneously learn absolute and relative notions of quality, enabling denser and more effective utilization of supervision signals from generated results of varying quality.
At inference, \(\mu_c^d\) is used as the dimension-wise reward, and the overall reward is computed as \(r_c=\frac{1}{|\mathcal{D}|}\sum_{d\in\mathcal{D}}\mu_c^d\). The score heads are used only during training and introduce no additional overhead for candidate ranking or online RFT.

\subsection{Foveation Calibration Objective}
Reliable VTON evaluation requires each reward dimension to attend to the appropriate visual evidence. 
However, due to broad visual-language alignment, VLMs typically lack attention to visual tokens and are highly distracted.
Therefore, we propose to use mask priors \cite{fashn-human-parser, xie2021segformer} to define a dimension-specific target token set \(\mathcal{S}_c^d\) for candidate \(c\): garment regions in the reference and generated images for garment consistency, person regions in the target and generated images for identity consistency, and the complete generated image for visual quality. These masks provide spatial supervision only during training and are not required at inference time.

Let \(a_{c,d,j}\) denote the attention assigned to visual token \(j\) when evaluating candidate \(c\) under dimension \(d\). The total attention mass within the corresponding target region is:
\begin{equation}
m_c^d
=
\sum_{j\in\mathcal{S}_c^d}
a_{c,d,j}.
\end{equation}
We encourage the model to concentrate on dimension-relevant visual evidence using
\begin{equation}
\mathcal{L}_{\mathrm{FCO}}
=
-\mathbb{E}_{c,d}
\left[
\log\left(m_c^d+\epsilon\right)
\right].
\end{equation}
This objective increases the attention allocated to relevant visual regions without imposing a fixed distribution within them, allowing the model to flexibly identify the evidence needed for each evaluation dimension. In practice, we compute FCO independently for each attention head and average the resulting losses.

\begin{table*}[ht]
\caption{Comparisons between various reward models in pairwise and K-way accuracies according to human preference.}
\label{tab:cloth2model_model2model}
\centering
\footnotesize
\renewcommand{\arraystretch}{1.18}
\setlength{\tabcolsep}{2.6pt}
\resizebox{\textwidth}{!}{
\begin{tabular}{l|cccc|cccccc|cccc|cccccc}
\toprule
\multirow{3}{*}{Method}
& \multicolumn{10}{c|}{{Cloth2Model}}
& \multicolumn{10}{c}{{Model2Model}} \\
\cline{2-21}
& \multicolumn{4}{c|}{Pairwise Acc.}
& \multicolumn{6}{c|}{K-way Acc.}
& \multicolumn{4}{c|}{Pairwise Acc.}
& \multicolumn{6}{c}{K-way Acc.} \\
\cline{2-21}
& \rule[-1.2ex]{0pt}{4.8ex} \makecell{Gar.\\Consist.}
& \makecell{Identity\\Consist.}
& \makecell{Visual\\Quality}
& \makecell{Overall\\Pair.}
& K=2 & K=3 & K=4 & K=5 & K=6 & Avg.
& \makecell{Gar.\\Consist.}
& \makecell{Identity\\Consist.}
& \makecell{Visual\\Quality}
& \makecell{Overall\\Pair.}
& K=2 & K=3 & K=4 & K=5 & K=6 & Avg. \\[6pt]
\hline

\multicolumn{1}{l|}{\textit{\textbf{Open-source Models}}} & \multicolumn{4}{c|}{} & \multicolumn{6}{c|}{} & \multicolumn{4}{c|}{} & \multicolumn{6}{c}{} \\
Qwen3-VL-4B
& 59.01 & 52.05 & 36.18 & 58.31
& 30.62 & 25.06 & 11.93 & 7.17 & 2.46 & 15.45
& 56.40 & 35.20 & 56.60 & 41.38
& 40.17 & 16.64 & 5.87 & 0.71 & 0.22 & 12.72 \\
Qwen3-VL-8B
& 63.49 & 51.19 & 39.26 & 52.20 & 74.53 & 14.90 & 5.42 & 4.04 & 1.79 & 20.14
& 62.38 & 50.52 & 58.01 & 52.22 & 56.63 & 19.96 & 6.82 & 1.78 & 0.65 & 17.16 \\
Qwen3-VL-32B
& 63.94 & 54.94 & 40.71 & 52.29 & \underline{76.96} & 23.25 & 8.24 & 5.61 & 2.91 & 23.39
& 63.18 & 53.21 & 57.92 & 55.55 & 60.58 & 23.41 & 7.18 & 2.20 & 1.30 & 18.93 \\
Qwen3.5-9B
& 62.20 & 49.61 & 41.15 & 49.84 & 63.96 & 16.48 & 5.64 & 3.81 & 2.24 & 18.43
& 61.28 & 44.63 & 57.59 & 50.70 & 52.62 & 19.45 & 8.47 & 3.25 & 1.41 & 17.04 \\
Qwen3.5-35B-A3B
& 64.34 & 49.84 & 31.58 & 49.64 & 73.17 & 12.64 & 3.47 & 1.57 & 0.67 & 18.30
& 63.00 & 60.49 & 52.56 & 60.64 & 56.69 & 17.41 & 6.90 & 2.79 & 1.94 & 17.15 \\
InternVL3.5-30B-A3B
& 58.15 & 50.60 & 37.65 & 48.93 & 65.58 & 18.51 & 4.77 & 2.47 & 1.12 & 18.49
& 58.83 & 43.99 & 55.66 & 50.10 & 55.52 & 16.60 & 7.58 & 1.62 & 1.95 & 16.65 \\

\hline
\multicolumn{1}{l|}{\textit{\textbf{Large-scale  Models}}} & \multicolumn{4}{c|}{} & \multicolumn{6}{c|}{} & \multicolumn{4}{c|}{} & \multicolumn{6}{c}{} \\
Qwen3.5-397B-A3B
& 66.81 & \underline{68.57} & 46.08 & 60.17
& 67.75 & 18.11 & \underline{12.94} & 6.25 & 0.00 & 21.01
& 63.04 & 73.79 & 58.10 & 63.07
& 63.98 & 22.99 & 8.60 & 5.26 & 0.00 & 20.17 \\
Gemini-3.0-Pro
& \textbf{69.10} & \underline{68.57}
& \underline{58.85} & \underline{64.50}
& 75.07 & \underline{26.86} & 9.76 & 6.50 & 2.91 & 24.22
& \underline{63.54} & \underline{75.24}
& 59.41 & \underline{67.28}
& \underline{69.93} & 31.73 & \underline{15.34}
& \underline{7.83} & \underline{3.88} & \underline{25.74} \\    
GPT-5.4
& 67.76 & 66.90 & 51.07 & 59.29 & 71.27 & 26.41 & 11.71 & \underline{8.74} & \underline{4.25} & \underline{24.48}
& 61.85 & 73.25 & \underline{59.62} & 65.36 & 67.08 & \underline{32.05} & 15.29 & 7.38 & 2.38 & 24.83 \\

\hline
\multicolumn{1}{l|}{\textit{\textbf{Reward Models (Ours)}}} & \multicolumn{4}{c|}{} & \multicolumn{6}{c|}{} & \multicolumn{4}{c|}{} & \multicolumn{6}{c}{} \\
TryOnReward (Ours)
& \underline{68.55} & \textbf{80.06} & \textbf{65.45} & \textbf{74.61}
& \textbf{80.49} & \textbf{58.47} & \textbf{45.64} & \textbf{21.40} & \textbf{12.04} & \textbf{43.61}
& \textbf{67.31} & \textbf{83.13} & \textbf{62.20} & \textbf{73.39}
& \textbf{82.50} & \textbf{60.79} & \textbf{42.52} & \textbf{27.23} & \textbf{10.02} & \textbf{44.62} \\
\hhline{-|----|------|----|------}
{$\Delta$ vs. Qwen3-VL-8B}
& {+5.06} & {+28.87} & {+26.19} & {+22.41}
& {+5.96} & {+43.57} & {+40.22} & {+17.36} & {+10.25} & {+23.47}
& {+4.93} & {+32.61} & {+4.19} & {+21.17}
& {+25.87} & {+40.83} & {+35.70} & {+25.45} & {+9.37} & {+27.46} \\
\bottomrule
\end{tabular}
}
\end{table*}

\begin{table*}[t]
    \caption{Results after reinforcement
fine-tuning using different reward models on the TryOn-Bench are listed for comparisons in the Cloth2Model and Model2Model scenarios. }
    \label{tab:rl_tryon_bench}
    \centering
    \small
    \setlength{\tabcolsep}{3.5pt}
    \resizebox{\textwidth}{!}{
        \begin{tabular}{llcccccccc}
            \toprule
            \multirow{2}{*}{Policy Model}
            & \multirow{2}{*}{Reward Model}
            & \multicolumn{4}{c}{Cloth2Model (product-display)}
            & \multicolumn{4}{c}{Model2Model (worn-on)} \\
            \cmidrule(lr){3-6}
            \cmidrule(lr){7-10}
            & &
            Gar. Consist. $\uparrow$
            & Identity Consist. $\uparrow$
            & Visual Quality $\uparrow$
            & Overall $\uparrow$
            & Gar. Consist. $\uparrow$
            & Identity Consist. $\uparrow$
            & Visual Quality $\uparrow$
            & Overall $\uparrow$ \\
            \midrule

            \multirow{5}{*}{Qwen-Image-Edit-2511}
            & \textit{Base}
            & 2.845 & 3.939 & 3.739 & 3.508
            & 2.471 & 3.354 & 3.316 & 3.047 \\

            & HPSv3
            & 2.694 & 3.407 & 3.717 & 3.273 
            & 2.421 & 2.486 & \underline{3.663} & 2.857 \\

            & EditReward
            & 2.906 & 3.967 & 3.676 & 3.517
            & 2.867 & {3.586} & 3.639 & 3.364 \\

            & Gemini-3.0-Pro
            & \underline{3.050}
            & \underline{3.991}
            & \underline{3.761}
            & \underline{3.600}
            & \underline{2.917}
            &  \underline{3.919}
            & {3.625}
            & \underline{3.519} \\

            & \textbf{TryOnReward (Ours)}
            & \textbf{3.137}
            & \textbf{4.015}
            & \textbf{3.804}
            & \textbf{3.652}
            & \textbf{3.010}
            & \textbf{3.921}
            & \textbf{3.760}
            & \textbf{3.564} \\

            \midrule

            \multirow{5}{*}{Qwen-Image-Edit-2511-SFT}
            & \textit{Base}
            & 2.878
            & 3.961
            & 3.796
            & 3.545
            & 2.629
            & 3.611
            & 3.708
            & 3.316 \\
            
            & HPSv3
            & 3.076
            & 4.008
            & 3.830
            & 3.638
            & 2.784
            & 3.435
            & 3.723
            & 3.314 \\
            
            & EditReward
            & 2.963
            & 4.010
            & 3.771
            & 3.582
            & 2.885
            & 3.423
            & 3.651
            & 3.319 \\
            
            & Gemini-3.0-Pro
            & \underline{3.110}
            & \underline{4.013}
            & \underline{3.796}
            & \underline{3.640}
            & \underline{2.909}
            & \underline{3.881}
            & \underline{3.698}
            & \underline{3.496} \\
            
            & \textbf{TryOnReward (Ours)}
            & \textbf{3.183}
            & \textbf{4.067}
            & \textbf{3.861}
            & \textbf{3.704}
            & \textbf{3.042}
            & \textbf{3.960}
            & \textbf{3.776}
            & \textbf{3.593} \\

            \bottomrule
        \end{tabular}
    }
\end{table*}
\subsection{Multi-Dimensional Pairwise Ranking}
Based on the uncertainty-aware reward formulation in Eq. \ref{eq:head}, we cast pairwise preference learning as probabilistic ranking instead of directly comparing deterministic reward scores. For each evaluation dimension \(d\), let \(w_d,l_d\in\mathcal{C}\) denote the preferred and rejected candidate indices, respectively. The dimension-wise preference probability is obtained by marginalizing the pairwise likelihood over the two latent reward distributions:
{
\begin{equation}
\begin{aligned}
P_d(I^{w_d} \succ I^{l_d})
&=
\int\!\!\int
\operatorname{sigmoid}(r_w-r_l) \\
&\quad \cdot
p_{w_d}(r_w)\,
p_{l_d}(r_l)\,
\mathrm{d}r_w\,\mathrm{d}r_l .
\end{aligned}
\label{eq:preference_probability}
\end{equation}
}
\noindent Here, \(p_{w_d}(r_w)\) and \(p_{l_d}(r_l)\) denote the Gaussian reward distributions predicted for the two candidates under dimension \(d\). By marginalizing over these distributions, the preference probability incorporates both the expected reward difference and the uncertainty associated with each prediction. The uncertainty-aware ranking loss is defined as the negative log-likelihood of the observed human preference:
\begin{equation}
\mathcal{L}_{\mathrm{rank}}^d
=
-\log P_d(I^{w_d} \succ I^{l_d}).
\label{eq:dimension_ranking_loss}
\end{equation}
In this way, the objective preserves dimension-specific preference signals while explicitly accounting for the uncertainty inherent in fine-grained VTON evaluation.

\subsection{Margin-Aware Pointwise Regression}
Pairwise preference supervision specifies only which candidate is preferred, without explicitly capturing their absolute quality levels or the magnitude of the preference. To recover this information, we use the score head to predict the human rating of each candidate. For candidate \(c\in\{A,B\}\), the score logits define a categorical distribution over the five rating levels:
\begin{equation}
p_c^d(k)
=
\operatorname{softmax}
\left(
\mathbf{z}_c^d
\right)_k, 
k\in\{1,\ldots,5\}.
\end{equation}
We first supervise the predicted rating distribution using a standard cross-entropy loss:
\begin{equation}
\mathcal{L}_{\mathrm{score}}^d
=
-\sum_{k=1}^{5}
y_{c,k}^d\log p_c^d(k).
\end{equation}
Here, \(y_{c,k}^d\) is the human rating label of candidate \(c\) under dimension \(d\). This objective anchors the learned representation to the semantics of the 1--5 rating scale, enabling the model to distinguish different absolute quality levels.

Additionally, since cross-entropy does not capture the ordinal structure of the ratings, we derive a differentiable expected score:
\begin{equation}
\widehat{s}_c^d
=
\sum_{k=1}^{5}
k\,p_c^d(k).
\end{equation}
Given two candidates with human ratings \(s_A^d\) and \(s_B^d\), we define \(\Delta_s=s_{\max}-s_{\min}\) as the maximum score difference. The normalized residual between predicted and annotated score gaps is:
\begin{equation}
e^d
=
\frac{
(\widehat{s}_A^d-\widehat{s}_B^d)
-
(s_A^d-s_B^d)
}{\Delta_s}.
\end{equation}
The margin-aware score-gap objective is formulated using Smooth L1 loss:
\begin{equation}
\mathcal{L}_{\mathrm{gap}}^d
=
\begin{cases}
\frac{1}{2}(e^d)^2,
& |e^d|<1, \\
|e^d|-\frac{1}{2},
& |e^d|\geq 1.
\end{cases}
\end{equation}
The annotated score gap provides an adaptive margin proportional to the preference strength. 
Through the shared VLM backbone, pointwise rating supervision and score-gap regression complement pairwise preference learning.

Combining all supervision signals, the overall training objective is:
{\small
\begin{equation}
\begin{aligned}
\mathcal{L}_{\mathrm{total}}
={}&
\frac{1}{|\mathcal{D}|}
\sum_{d\in\mathcal{D}}
\left(\mathcal{L}_{\mathrm{rank}}
+
\lambda_{\mathrm{score}}\mathcal{L}_{\mathrm{score}}^d
+
\lambda_{\mathrm{gap}}\mathcal{L}_{\mathrm{gap}}^d
\right) \\
&+
\lambda_{\mathrm{FCO}}\mathcal{L}_{\mathrm{FCO}}.
\end{aligned}
\label{eq:total_loss}
\end{equation}
}
The ranking objective learns relative preferences, the score objectives preserve absolute rating semantics and preference strength, and Foveation Calibration Objective grounds the resulting rewards in dimension-relevant visual evidence.

\begin{table*}[t]
    \caption{Quantitative comparisons with the state-of-the-art try-on methods on the DressCode and VITON-HD benchmarks.}
    \label{tab:vton_comparison}
    \centering
    \footnotesize
    \setlength{\tabcolsep}{5.5pt}
    \renewcommand{\arraystretch}{1.0}
    \resizebox{\textwidth}{!}{
        \begin{tabular}{lcccccccccccc}
            \toprule
            \multirow{3}{*}{Method}
            & \multicolumn{6}{c}{DressCode}
            & \multicolumn{6}{c}{VITON-HD} \\
            \cmidrule(lr){2-7}
            \cmidrule(lr){8-13}

            & \multicolumn{4}{c}{Paired}
            & \multicolumn{2}{c}{Unpaired}
            & \multicolumn{4}{c}{Paired}
            & \multicolumn{2}{c}{Unpaired} \\
            \cmidrule(lr){2-5}
            \cmidrule(lr){6-7}
            \cmidrule(lr){8-11}
            \cmidrule(lr){12-13}

            & FID $\downarrow$
            & KID $\downarrow$
            & SSIM $\uparrow$
            & LPIPS $\downarrow$
            & FID $\downarrow$
            & KID $\downarrow$
            & FID $\downarrow$
            & KID $\downarrow$
            & SSIM $\uparrow$
            & LPIPS $\downarrow$
            & FID $\downarrow$
            & KID $\downarrow$ \\
            \midrule

            IDM-VTON
            & 7.181 & 3.524 & 0.891 & 0.070 & 9.167 & 4.489
            & 6.112 & 1.112 & 0.866 & 0.074 & 9.249 & 1.267 \\

            OOTDiffusion
            & 6.975 & 2.014 & 0.873 & 0.077 & 8.121 & 2.886
            & 5.762 & 1.267 & 0.843 & \underline{0.072}
            & 9.082 & 0.702 \\

            CatVTON
            & 5.031 & 1.789 & \underline{0.899} & 0.074
            & 7.485 & 2.717
            & 8.055 & 1.659 & \underline{0.888} & 0.076
            & 11.404 & 2.374 \\

            Leffa
            & 7.193 & 2.114 & 0.861 & 0.084 & 20.099 & 13.506
            & \underline{5.743} & 0.692 & 0.857 & 0.076
            & 10.446 & 2.640 \\

            PromptDresser
            & 9.563 & 4.795 & 0.858 & 0.104 & 10.618 & 4.978
            & 5.934 & \underline{0.550} & 0.846 & 0.090
            & \underline{8.885} & \underline{0.909} \\

            Any2AnyTryon
            & 4.893 &
            1.252 &
            0.881 &
            0.082 &
            6.540 &
            1.667 &
            11.532 &
            3.884 &
            0.875 &
            0.129 &
            14.756 &
            5.084 \\

            FastFit
            & \underline{3.658} & 
            \textbf{0.747} & 
            0.889 & 
            \underline{0.070} & 
            \textbf{5.357} & 
            \textbf{1.027} & 
            
            6.863 & 
            0.949 & 
            0.865 & 
            0.096 & 
            9.491 & 
            0.973 \\
            
            \midrule
            {Gemini-3.0-Pro-RFT}
            & 5.095 & 0.900 & 0.868 & 0.112
            & 6.179 & 1.573
            & 5.978 & 1.191
            & 0.897 & 0.088
            & 10.119 
            & 2.132 \\
            \textbf{TryOnReward-RFT}
            & \textbf{2.877} & \underline{0.876} & \textbf{0.927} & \textbf{0.063}
            & \underline{5.611} & \underline{1.319}
            & \textbf{5.723} & \textbf{0.533}
            & \textbf{0.906} & \textbf{0.061}
            & \textbf{7.652} 
            & \textbf{0.710} \\

            \bottomrule
        \end{tabular}
    }

\end{table*}

\begin{table*}[t]
\caption{Ablation study of TryOnReward on TryOnRewardBench. The last two rows remove FCO alone and FCO together with pointwise supervision, respectively.}
\label{tab:ablation}
\centering
\footnotesize
\renewcommand{\arraystretch}{1.18}
\setlength{\tabcolsep}{2.6pt}
\resizebox{\textwidth}{!}{
\begin{tabular}{l|cccc|cccccc|cccc|cccccc}
\toprule
\multirow{3}{*}{Models}
& \multicolumn{10}{c|}{Cloth2Model}
& \multicolumn{10}{c}{Model2Model} \\
\cline{2-21}
& \multicolumn{4}{c|}{Pairwise Acc.}
& \multicolumn{6}{c|}{K-way Acc.}
& \multicolumn{4}{c|}{Pairwise Acc.}
& \multicolumn{6}{c}{K-way Acc.} \\
\cline{2-21}
& \rule[-1.2ex]{0pt}{4.8ex}\makecell{Gar.\\Consist.}
& \makecell{Identity\\Consist.}
& \makecell{Visual\\Quality}
& \makecell{Overall\\Pair.}
& K=2 & K=3 & K=4 & K=5 & K=6 & Avg.
& \makecell{Gar.\\Consist.}
& \makecell{Identity\\Consist.}
& \makecell{Visual\\Quality}
& \makecell{Overall\\Pair.}
& K=2 & K=3 & K=4 & K=5 & K=6 & Avg. \\[6pt]
\hline

\textbf{TryOnReward (Full)}
& \textbf{68.55} & \textbf{80.06} & 65.45 & \textbf{74.61}
& 80.49 & \textbf{58.47} & \textbf{45.64}
& 21.40 & \underline{12.04} & \textbf{43.61}
& \textbf{67.31} & \textbf{83.13} & 62.20 & \textbf{73.39}
& \textbf{82.50} & \textbf{60.79} & \textbf{42.52}
& \textbf{27.23} & \textbf{10.02} & \textbf{44.62} \\

- FCO
& 67.46 & \underline{79.24} & \underline{68.93} & \underline{74.10}
& \textbf{82.93} & \underline{50.56} & \underline{40.37}
& \textbf{23.20} & \textbf{12.50} & \underline{41.91}
& 65.15 & 81.25 & \textbf{64.44} & \underline{71.72}
& \underline{82.03} & 52.01 & \underline{34.33}
& \underline{23.08} & 8.59 & \underline{40.01} \\

\makecell[l]{- Pointwise Supervision}
& \underline{67.47} & 78.13 & \textbf{69.49} & 72.81
& \underline{82.11} & 50.34 & 39.91
& \underline{21.62} & 11.57 & 41.11
& \underline{65.60} & \underline{82.94} & \underline{64.24} & 70.62
& 80.10 & \underline{52.44} & 31.76
& 21.44 & \underline{9.36} & 39.02 \\

\bottomrule
\end{tabular}
}
\end{table*}

\section{Experiments}
\subsection{Implementation Details}
We adopt Qwen3VL-8B-Instruct \cite{bai2025qwen3} as the pretrained backbone and perform full-parameter fine-tuning to optimize the entire model. The learning rate is \(2 \times 10^{-6}\) with a cosine learning rate schedule and a warm-up ratio of 0.05. The total batch size is 16. To improve training stability, we preprocess all input images to a unified resolution of \(672 \times 672\), while preserving their original aspect ratios during resizing. We set the loss weights $\lambda_{\mathrm{score}}$, $\lambda_{\mathrm{gap}}$, and $\lambda_{\mathrm{FCO}}$ to 0.1, 0.05, and 0.05, respectively.

\subsection{Evaluation Setting} 
We evaluate TryOnReward in terms of human-preference alignment and its effectiveness as a reward for online RFT. The reported quantitative results are averaged over multiple random seeds. For preference alignment, we measure the agreement between model predictions and human annotations on TryOnRewardBench. The scoring dimensions of existing general-purpose text-to-image and image-editing reward models do not directly match our VTON-specific criteria, and dedicated VTON reward baselines are unavailable. We therefore compare against general-purpose VLMs across different model scales. Our open-source baselines include Qwen3-VL (4B, 8B, and 32B)~\cite{bai2025qwen3}, Qwen3.5 (9B and 35B-A3B)~\cite{qwen3.5}, and InternVL3.5-30B-A3B~\cite{wang2025internvl3}. We further include large-scale baselines comprising Qwen3.5-397B~\cite{qwen3.5} and proprietary models GPT-5.4~\cite{openai_gpt54} and Gemini-3.0-Pro~\cite{pichai2025new}. We report pairwise and strict \(K\)-way accuracy. Pairwise accuracy measures agreement between predicted and human preferences. For overall accuracy, the GT preference is determined by the mean human rating across garment consistency, identity consistency, and visual quality and compared with that derived from the predicted overall scores. For \(K\)-way accuracy, following a fixed generator ordering, we rank the first \(K\) results for each input by their predicted overall scores and compare them with the human ranking derived from mean dimension-wise ratings. Thus, \(K=2\) evaluates the designated first two generators. A ranking is correct only if all \(\mathrm{C}_K^2\) predicted pairwise relations match the human ranking. We report results for \(K=2,\ldots,6\).

    
    
    

For reward-guided RFT, we conduct online post-training using DiffusionNFT~\cite{zheng2025diffusionnft} and evaluate the resulting policies on TryOn-Bench. We adopt Qwen-Image-Edit-2511~\cite{wu2025qwen} and a fine-tuned VTON-SFT version as policy models, and compare with HPSv3~\cite{ma2025hpsv3}, EditReward~\cite{wu2025editreward}, and Gemini-3.0-Pro with rubric-based scoring as reward functions.
We use GPT-5 as an independent automatic evaluator, rating each result from 1 to 5 in Garment Consistency, Identity Consistency, and Visual Quality, with their average as the Overall Score. This choice is motivated by our evaluator assessment, which indicates that GPT-5's judgments on post-RFT outputs are comparatively aligned with human assessments. GPT-5 is distinct from all compared reward functions, and the same evaluation prompt and scoring protocol are applied uniformly across methods.

We additionally evaluate the RFT-optimized models on DressCode~\cite{morelli2022dress} and VITON-HD~\cite{choi2021viton}. We report results under both paired and unpaired settings~\cite{chong2025fastfitacceleratingmultireferencevirtual}. In the paired setting, the target garment is the one worn by the person in the original dataset image, which serves as the ground truth for reconstruction; we report FID, KID, LPIPS, and SSIM. In the unpaired setting, the person is matched with a different garment, and no corresponding ground-truth try-on image is available; we report FID and KID to compare the distributions of generated and real images. Our models and all baselines are evaluated under the same paired/unpaired protocol. We compare against representative VTON methods, including IDM-VTON~\cite{choi2024idmvton}, OOTDiffusion~\cite{xu2025ootdiffusion}, CatVTON~\cite{chong2025catvton}, Leffa~\cite{zhou2025learning}, PromptDresser~\cite{kim2024promptdresser}, Any2AnyTryon~\cite{guo2025any2anytryon}, and FastFit~\cite{chong2025fastfitacceleratingmultireferencevirtual}, to assess their performance on established VTON benchmarks.

\begin{figure}[t]
    \centering
    \includegraphics[width=\linewidth]{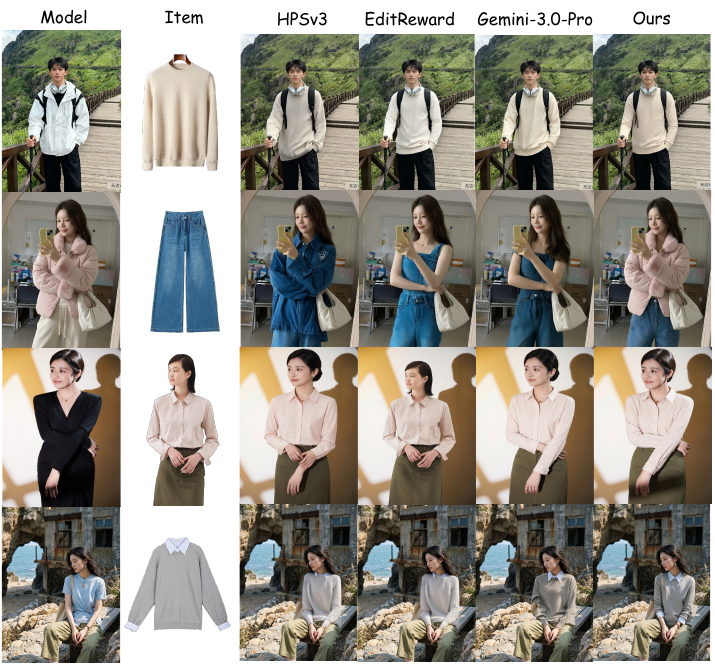}   
    
    \caption{Qualitative comparisons of try-on results after RFT with different reward models.}
    \label{fig:tryon_case}
\end{figure}

\subsection{Main Results}
\textbf{Human preference alignment:}
As shown in Table~\ref{tab:cloth2model_model2model}, TryOnReward achieves the best overall pairwise accuracy on both Cloth2Model and Model2Model scenarios, reaching 74.61\% and 73.39\%, respectively. It substantially outperforms open-source VLM baselines as well as proprietary models such as GPT-5.4 and Gemini-3.0-Pro. 
Notably, TryOnReward also shows clear advantages under the more stringent \(K\)-way Accuracy, with average \(K\)-way scores of 43.61\% and 44.62\% on the two subsets, far exceeding the strongest proprietary baselines. These results demonstrate that VTON-specific multi-dimensional reward modeling better aligns with human preferences. They also indicate that the model can not only effectively discriminate pairwise preferences but also support more stable global ranking among multiple candidates.

\noindent
\textbf{Results for RFT training:}
As shown in Tab.~\ref{tab:rl_tryon_bench}, TryOnReward improves both policy models on TryOn-Bench. The gain is particularly pronounced for Qwen-Image-Edit-2511 on Model2Model, where the overall score increases from 3.047 to 3.564, demonstrating the effectiveness of our model for challenging cross-person try-on. We further evaluate the optimized model on the DressCode and VITON-HD benchmarks. As shown in Tab.~\ref{tab:vton_comparison}, the model achieves competitive performance compared with representative VTON methods.

\begin{figure}[t]
    \centering
    \includegraphics[width=\linewidth]{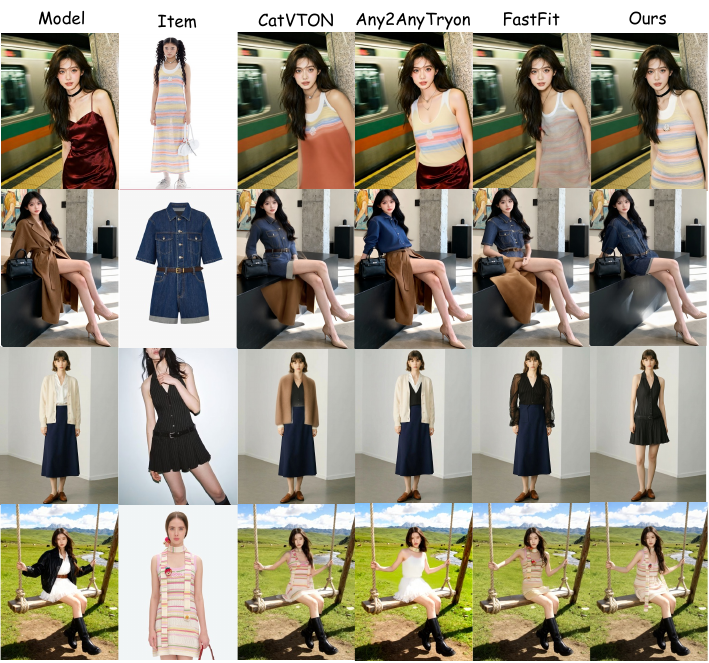}   
    
    \caption{Qualitative comparisons across virtual try-on methods using flat-lay and worn garment references.}
    \label{fig:tryon_case2}
    
\end{figure}

\subsection{Ablations}
Tab. \ref{tab:ablation} presents a component-removal ablation study of the proposed method. Removing FCO from the full model reduces overall pairwise and average \(K\)-way accuracy in both Cloth2Model and Model2Model, with the largest declines observed in garment and identity consistency, highlighting the benefit of localized cross-image grounding. Further removing pointwise supervision leads to an additional drop in overall alignment and ranking performance, confirming its complementary role in capturing absolute quality semantics and preference strength beyond pairwise comparisons. Visual Quality accuracy, however, increases slightly when FCO is removed, likely because its holistic nature makes the full-image mask a coarse and overly restrictive prior.
\subsection{Qualitative Results}
Fig.~\ref{fig:tryon_case} compares try-on results obtained after RFT with different reward models. In the first two rows, the policy optimized with TryOnReward more faithfully preserves garment appearance and structure while avoiding unintended changes to non-target clothing. It also better retains the target person's facial features and original pose in the third row, and produces sharper textures and clearer local details in the fourth row, where baseline outputs exhibit noticeable blur. These qualitative observations are consistent with the quantitative results reported in Table~\ref{tab:rl_tryon_bench}. Fig.~\ref{fig:tryon_case2} further shows that our policy better preserves garment patterns and structure, as well as person identity and pose, compared with representative VTON methods using flat-lay and worn garment references. The attention maps in Fig.~\ref{fig:motivation} further illustrate the differences in visual grounding: general-purpose VLMs exhibit diffuse attention and are frequently distracted by task-irrelevant regions, while TryOnReward with FCO attends more closely to dimension-relevant regions and their cross-image correspondences, providing more reliable visual evidence for fine-grained VTON evaluation.

\section{Conclusion}
In this paper, we present {TryOnReward}, a fine-grained reward model tailored for virtual try-on. TryOnReward decomposes human preferences into garment consistency, identity consistency, and visual quality, and introduces a dimension-specific foveation calibration learning approach to ground reward predictions in task-relevant visual evidence. It further unifies pair-wise preference learning and point-wise score regression, enabling the learned reward space to capture ranking consistency, rating semantics, and preference strength. To support training and systematic evaluation of reward models, we construct {TryOnRank-100K}, TryOn-Bench and {TryOnRewardBench}. Extensive experiments demonstrate that TryOnReward can achieve better alignment with human preferences and provide effective reward signals for reinforcement learning with try-on generative models.

{\fontsize{9}{10}\selectfont
\setlength{\bibsep}{0pt}
\bibliographystyle{ieeenat_fullname}
\bibliography{main}
}

\end{document}